\documentclass[10pt,twocolumn,letterpaper]{article}

\usepackage{wacv}              

\usepackage[accsupp]{axessibility}  
\usepackage[utf8]{inputenc} 
\usepackage[T1]{fontenc}    
\usepackage{tabularx}
\usepackage{adjustbox}
\usepackage{pgfplots}
\pgfplotsset{compat=1.18}

\definecolor{easyblue}{RGB}{33,150,243}
\definecolor{hardorange}{RGB}{255,112,67}
\definecolor{exp1}{HTML}{FF6F61}
\definecolor{exp2}{HTML}{FFB74D}
\definecolor{exp3}{HTML}{4DB6AC}
\definecolor{exp4}{HTML}{4FC3F7}

\newcolumntype{Y}{>{\centering\arraybackslash}X}
\newcolumntype{Z}{>{\columncolor{gray!10}}Y}
\newcolumntype{g}{>{\columncolor{gray!10}}c}

\newcommand{\plotExperiments}[3]{
\begin{tikzpicture}
    \begin{axis}[
        scale only axis,
        xmin=0, xmax=31,
        ymin=28,
        ymax=75,
        legend style={at={(0.5,-0.2)}, anchor=north, legend columns=-1, font=\scriptsize, line width=0.3pt,draw=none},
        width=0.9\linewidth,
        height=2.2cm,
        xtick={0, 10, 20, 30},
        ytick={30, 40, 50, 60},
        xticklabel style={font=\scriptsize},
        yticklabel style={font=\scriptsize},
        xlabel style={font=\scriptsize},
        ylabel style={font=\scriptsize},
        axis x line*=bottom,
        axis y line*=left,
        axis line style={-latex},
    ]
    
    \fill[exp2!10] (axis cs:0.1,28) rectangle (axis cs:10,70);
    \fill[exp3!10] (axis cs:10,28) rectangle (axis cs:20,70);
    \fill[exp4!10] (axis cs:20,28) rectangle (axis cs:30,70);
    
    \draw[dashed, black!50] (axis cs:10,28) -- (axis cs:10,75);
    \draw[dashed, black!50] (axis cs:20,28) -- (axis cs:20,75);
    \draw[dashed, black!50] (axis cs:30,28) -- (axis cs:30,75);

    \node[anchor=south,font=\scriptsize] at (rel axis cs:0.15,0.86) {LL};
    \node[anchor=south,font=\scriptsize] at (rel axis cs:0.48,0.86) {VLL};
    \node[anchor=south,font=\scriptsize] at (rel axis cs:0.82,0.86) {ELL};

    {#1}
    {#2}
    {#3}
    
    \end{axis}
\end{tikzpicture}
\vspace{-0.01pt}
}

\newcommand{\plotExperimentCustom}[5]{ 
    \plotExperiments{
        \addplot[color=exp1, thick] coordinates {#1};
        \addplot[color=exp2, thick] coordinates {#2};
        \addplot[color=exp3, thick] coordinates {#3};
        \addplot[color=exp4, thick] coordinates {#4};
    }{}{#5}
}

\newcommand{\plotDefaultLegend}{
    \addlegendimage{line legend, thick, color=exp1}
    \addlegendentry{WL AP}
    
    \addlegendimage{line legend, thick, color=exp2}
    \addlegendentry{LL AP}
    
    \addlegendimage{line legend, thick, color=exp3}
    \addlegendentry{VLL AP}
    
    \addlegendimage{line legend, thick, color=exp4}
    \addlegendentry{ELL AP}
}

\newcommand{\plotExperiment}[4]{
    \plotExperimentCustom{#1}{#2}{#3}{#4}{
        \plotDefaultLegend
    }
}

\definecolor{wacvblue}{rgb}{0.21,0.49,0.74}
\usepackage[pagebackref,breaklinks,colorlinks,allcolors=wacvblue]{hyperref}

\def\wacvPaperID{2519} 
\def\confName{WACV}
\def\confYear{2026}

\title{PoseAdapt: Sustainable Human Pose Estimation via Continual Learning Benchmarks and Toolkit}

\author{Muhammad Saif Ullah Khan \space and \space Didier Stricker\\
German Research Center for Artificial Intelligence (DFKI)\\
Trippstadter Str. 122, 67663 Kaiserslautern, DE\\
{\tt\small \url{https://saifkhichi96.github.io/research/poseadapt/}}
}

\begin{document}
\maketitle
\thispagestyle{empty}
Human pose estimators are typically retrained from scratch or naively fine‑tuned whenever keypoint sets, sensing modalities, or deployment domains change—an inefficient, compute‑intensive practice that rarely matches field constraints. We present \textbf{PoseAdapt}, an open-source framework and benchmark suite for \emph{continual pose model adaptation}. PoseAdapt defines domain‑incremental and class‑incremental tracks that simulate realistic changes in density, lighting, and sensing modality, as well as skeleton growth. The toolkit supports two workflows: (i) \emph{Strategy Benchmarking}, which lets researchers implement continual learning (CL) methods as plugins and evaluate them under standardized protocols; and (ii) \emph{Model Adaptation}, which allows practitioners to adapt strong pretrained models to new tasks with minimal supervision. We evaluate representative regularization-based methods in single‑step and sequential settings. Benchmarks enforce a fixed lightweight backbone, no access to past data, and tight per‑step budgets. This isolates adaptation strategy effects, highlighting the difficulty of maintaining accuracy under strict resource limits. PoseAdapt connects modern CL techniques with practical pose estimation needs, enabling adaptable models that improve over time without repeated full retraining.

\section{Introduction}
\label{sec:intro}

Human pose estimation enables applications across autonomous systems~\cite{guesdon2021dripe,zheng2022multi,wang2023learning}, healthcare~\cite{dibenedetto2024new,zhou2024gait,yin2025progait}, sports analytics~\cite{ingwersen2023sportspose,yeung2025athletepose3d}, and human–computer interaction~\cite{anvari20223d,huo20233d,son2025interaction}. Advances in representation methods~\cite{zhang2020distribution,wang2022contextual,li2022simcc,purkrabek2025probpose}, model architectures~\cite{maji2022yolo,xu2022vitpose,jiang2023rtmpose}, data scale~\cite{andriluka20142d,liu2020new,moon2020interhand,jin2020whole,khan2025towards}, and training pipelines~\cite{joo2021exemplar,marusov2022enhancing} have pushed 2D keypoint benchmarks toward saturation~\cite{gao2025systematic}.

\begin{figure}
    \centering
    \includegraphics[width=\linewidth]{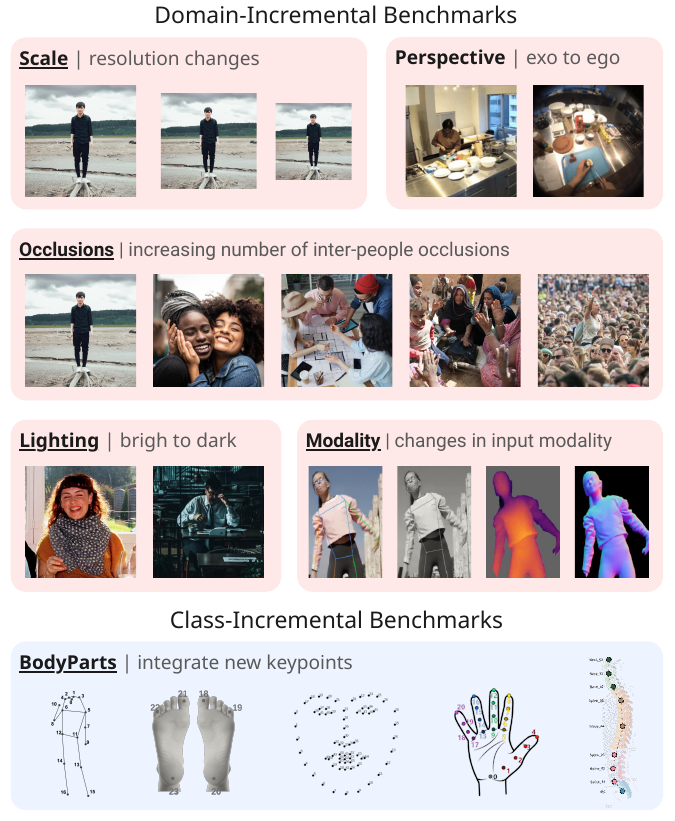}
    \caption{\textbf{PoseAdapt Benchmarks.}
    We introduce a diverse suite of domain- and class-incremental benchmarks for human pose estimation. \textit{Top}: Domain-incremental settings simulate increasing difficulty through scale, perspective, occlusion, lighting, and modality shifts. \textit{Bottom}: Class-incremental benchmarks gradually add new keypoint types to evaluate the ability to extend skeletons over time. All benchmarks share a fixed backbone, fixed per-step data budget, and unified evaluation protocol.}
    \label{fig:benchmarks}
\end{figure}

\begin{figure}[th]
\centering
\begin{tikzpicture}
  \begin{axis}[
    ybar, axis on top,
    height=3.5cm, width=\linewidth,
    bar width=0.3cm,
    tick align=inside,
    ymin=0, ymax=80,
    enlarge x limits=0.2,
    legend style={
        draw=none,
        at={(0.5,-0.2)},
        anchor=north,
        legend columns=-1,
        font=\scriptsize,
        /tikz/every even column/.append style={column sep=0.1cm}
    },
    tick label style={font=\scriptsize},
    label style={font=\scriptsize},
    ylabel={AP},
    symbolic x coords={Density,Lighting,Modality},
    xtick=data,
    nodes near coords={},
  ]

  \addplot [draw=none, fill=blue!50] coordinates {
      (Density,70.06)
      (Lighting,70.06)
      (Modality,70.06)
  };

  \addplot [draw=none, fill=red!50] coordinates {
      (Density,64.01)
      (Lighting,65.33)
      (Modality,62.36)
  };

  \addplot [draw=none, fill=red!50] coordinates {
      (Density,51.04)
      (Lighting,57.72)
      (Modality,16.06)
  };

  \addplot [draw=none, fill=red!50] coordinates {
      (Density,51.10)
      (Lighting,48.74)
  };

  \legend{Reference Data,Adaptation Data}
  \end{axis}
\end{tikzpicture}
\begin{tikzpicture}
\begin{axis}[
    axis on top,
    xlabel={Brightness},
    ylabel={AP},
    width=0.65\linewidth, height=3cm,
    xmin=0.1, xmax=1.1, ymin=40.0, ymax=80.0,
    grid=both,
    tick label style={font=\scriptsize},
    label style={font=\scriptsize},
]
\addplot+[
    mark=o,
    color=easyblue,
    thick
] coordinates {
(1.0,70.06)
(0.8,65.22) 
(0.6,64.14)
(0.4,61.62)
(0.2,54.67)
};
\end{axis}
\end{tikzpicture}
\caption{\textbf{Off-the-shelf models struggle under realistic shifts.} \textit{Top:} Accuracy on the pretrained reference dataset (blue) drops consistently under sequential shifts in density, lighting, and modality (red), even though these changes are relatively minor compared to training conditions. \textit{Bottom}: When brightness is progressively reduced on the same dataset, AP declines steadily, underscoring the brittleness of static models to illumination variation.}
\label{fig:shift-problem}
\end{figure}

Despite these gains, state‑of‑the‑art models are fundamentally static: they are trained once on fixed datasets and deployed under the assumption that test distributions match training. In practice, changes in illumination, viewpoint, density, or sensing modality cause significant drops in accuracy~\cite{essich2023auxiliary,chellappa2024some}. These limitations are especially visible in dynamic or resource‑constrained settings (e.g., high‑speed sports~\cite{yeung2025athletepose3d,ingwersen2023sportspose} or egocentric capture~\cite{wang2022estimating,huang2024egoexolearn}), where strong motion, occlusion, and field‑of‑view shifts challenge standard training regimes. These domains often involve domain-specific skeletons or sensing modalities for which pretrained estimators are either mismatched or incomplete. As shown in \cref{fig:shift-problem}, off‑the‑shelf models degrade under realistic shifts, undermining their deployment-time performance despite high accuracy on benchmark datasets.

\begin{figure}[ht]
    \centering
    \includegraphics[width=\linewidth]{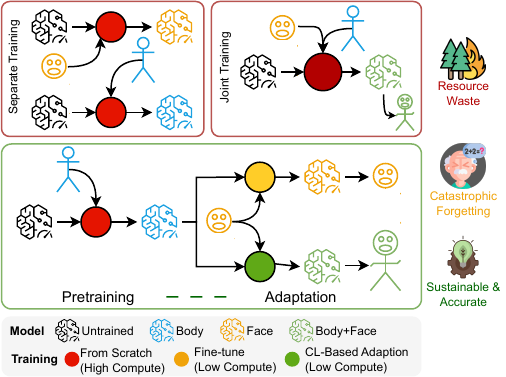}
    \caption{\textbf{Adaptation strategy comparison} \textit{Top}: Conventional solutions either train models separately (resource waste) or finetune (prone to forgetting). \textit{Bottom}: PoseAdapt enables adaptation using continual learning (CL) techniques to retain prior knowledge while specializing to new skeletons or domains.}
    \label{fig:adaptation}
\end{figure}

To overcome these deployment-time limitations or skeleton mismatches, off-the-shelf models often need to be fine-tuned on domain-specific data before they can be integrated in a real-world application. In \cref{fig:adaptation}, we compare different adaptation strategies used in practice. Retraining a model from scratch for each deployment condition is costly and slow, and naive fine‑tuning tends to overwrite prior knowledge, leading to catastrophic forgetting~\cite{kirkpatrick2017overcoming,li2017learning}. Some recent efforts target cross‑skeleton generalization~\cite{chen2025recurrent}, but typically rely on large backbones or extensive supervision, limiting deployability on edge devices. In contrast, continual learning (CL) methods regularize updates to preserve previously acquired knowledge while specializing to new experiences~\cite{kirkpatrick2017overcoming,li2017learning}. Therefore, \emph{we advocate continual adaptation as a sustainable alternative.} Instead of discarding prior competence, models should incrementally incorporate new domains or keypoints while retaining past performance.

To this end, \textbf{we propose PoseAdapt, which operationalizes the continual adaptation principle for human pose estimation}. This includes (1) an open-source framework for adapting pretrained estimators to domain-specific datasets using established CL strategies, and (2) a dedicated benchmark for comparing performance of different CL strategies under strict, deployment‑oriented constraints for both domain‑incremental and class‑incremental scenarios. Our PoseAdapt framework allows practitioners to trivially fine-tune off-the-shelf estimators on their application-specific data in a sustainable manner. Concurrently, the principled PoseAdapt benchmark encourages researchers to develop novel CL strategies for human pose estimation tasks by enabling fair and reproducible comparison.

The plug-and-play design of our PoseAdapt framework supports future extensions to new adaptation methods. This provides an evolving resource for human pose estimation practitioners and researchers which formalizes continual learning for human pose estimation.

\noindent \textbf{Contributions.}
\begin{enumerate}
    \item We introduce PoseAdapt, an open-source framework for continual learning in pose estimation, with support for domain- and class-incremental scenarios\footnote{Source code available at \url{https://github.com/dfki-av/poseadapt/}. Experiments reported in this paper use code from the ``\texttt{wacv-2026-camera-ready}'' branch.}.
    \item We design challenging, realistic benchmark protocols that capture gradual distribution shifts in resolution, occlusion, lighting, modality, and skeleton structure.
    \item We release modular toolkit with dataset wrappers, plugin-based CL strategies, and protocol-aware evaluation tools to facilitate sustainable pose model adaptation research.
\end{enumerate}

\section{Related Work}
\label{sec:related_work}

\noindent \textbf{Pose Estimation Toolkits and Distribution Shifts.}
Open-source systems such as OpenPose~\citep{cao2019openpose}, Detectron2~\citep{wu2019detectron2}, MMPose~\citep{mmpose2020}, AlphaPose~\citep{fang2022alphapose}, RTMPose~\citep{jiang2023rtmpose}, and ViTPose~\citep{xu2022vitpose} have driven rapid progress in 2D human pose estimation with modular training and evaluation across datasets and backbones. Despite their breadth, these toolkits largely assume a static train--test regime in which models are trained once on curated data and deployed unchanged. Related adaptation efforts in vision---for example, test-time adaptation via entropy minimization~\citep{wang2021tent}---offer robustness within a domain but do not address \emph{incremental} task streams, skeleton growth, or long-horizon retention. Some cross-skeleton methods jointly train on multiple datasets~\cite{sarandi2023learning}, which is inefficient and impractical for continual evolution. Consequently, deploying pose estimators in dynamic settings (wearables, robotics, sports analytics) remains brittle when illumination, viewpoint, density, or sensing modality shift over time. \textit{PoseAdapt framework provides continual adaptation protocols for pose estimation, missing in existing systems.}

\noindent \textbf{Benchmarks and Datasets for Pose Estimation.}
Large-scale benchmarks (COCO Keypoints~\citep{lin2014microsoft}, MPII~\citep{andriluka20142d}) and newer efforts (COCO-WholeBody~\citep{jin2020whole}, PoseTrack~\citep{andriluka2018posetrack}, Halpe~\citep{fang2022alphapose}, SpineTrack~\citep{khan2025towards}) have standardized evaluation and spurred architectural advances. However, these benchmarks are organized around static splits and assume full retraining when the task changes (e.g., skeleton extensions or cross-modality transfer). In contrast, continual learning (CL) benchmarks in classification study nonstationary streams and forgetting using protocols such as Split-CIFAR~\citep{zenke2017continual}, CORe50~\citep{lomonaco2017core50}, and DomainNet~\citep{peng2019moment}. \textit{An analogous, pose-specific benchmark that enforces realistic constraints (fixed lightweight backbones, no access to past data, limited budgets) has been lacking; PoseAdapt fills this gap with domain- and class-incremental tracks.}

\begin{figure*}[t]
    \begin{subfigure}[t]{0.50\linewidth}
        \centering
        \includegraphics[width=\linewidth]{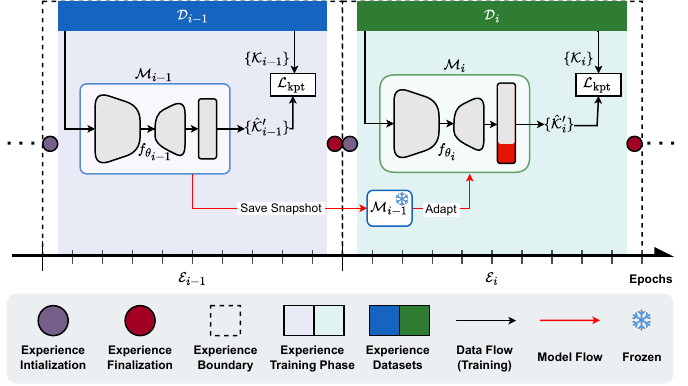}
    \end{subfigure}
    \hfill
    \begin{subfigure}[t]{0.45\linewidth}
        \centering
        \includegraphics[width=\linewidth]{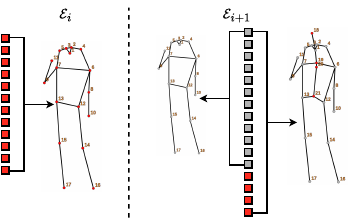}
    \end{subfigure}
    \caption{\textbf{PoseAdapt Framework.} 
    \emph{Left:} At each experience $\mathcal{E}_i$, PoseAdapt initializes the model for the new experience (e.g., snapshot creation, head expansion, or architecture-specific adjustments), followed by an adaptation phase that optimizes the model on $\mathcal{D}_i$ with strategy-defined regularization, and a finalization step to compute and store any statistics for later experiences. 
    \emph{Right:} Head expansion for class-incremental experiences, where the output dimensionality grows as new keypoints are introduced.}
    \label{fig:poseadapt-framework}
\end{figure*}

\noindent \textbf{Continual Learning: Paradigms and Constraints.}
CL aims to learn from nonstationary streams without catastrophic forgetting~\citep{mccloskey1989catastrophic,parisi2019continual,van2022three}. Representative approaches include: \emph{regularization-based} methods that constrain parameter or function drift (EWC~\citep{kirkpatrick2017overcoming}, SI~\citep{zenke2017continual}, LwF~\citep{li2017learning}, LFL~\cite{jung2016less}, IMM~\cite{lee2017overcoming}); \emph{replay-based} methods that use exemplars or generative rehearsal (iCaRL~\citep{rebuffi2017icarl}, DGR~\citep{shin2017continual}, A-GEM~\citep{chaudhry2019efficient}, DER~\citep{buzzega2020dark}); and \emph{architectural} approaches that isolate or grow capacity (PNNs~\citep{rusu2016progressive}, PackNet~\citep{mallya2018packnet}, Piggyback~\citep{mallya2018piggyback}, HAT~\citep{serra2018overcoming}). Many of these assume either access to a replay buffer or the ability to expand parameters and heads per task, which conflicts with deployment constraints such as fixed memory, strict compute, and privacy limits on past data. \textit{PoseAdapt benchmark isolates adaptation strategy effects by fixing model architecture, disallowing access to old data, and enforcing per-step budgets, enabling fair comparison under realistic constraints.}

\noindent \textbf{Continual Learning for Dense Prediction.}
Transferring CL to dense prediction introduces unique challenges: spatial outputs, structured losses, and label-evolution (e.g., new classes or keypoints). Recent work in segmentation~\citep{cermelli2020modeling,michieli2019incremental} and detection~\citep{shmelkov2017incremental} adapts regularization, distillation, or pseudo-labeling to retain prior categories while learning new ones. For human pose estimation, CL remains comparatively underexplored; early studies adapt CL ideas to structured outputs~\citep{gu2022not}, but evaluations are heterogeneous, often using small-scale or ad hoc protocols. \textit{By standardizing domain-incremental and class-incremental settings with shared budgets and metrics (forward transfer, retention), PoseAdapt provides the first controlled testbed to assess CL strategies for pose at scale.}

Beyond benchmarks, software ecosystems for CL such as Avalanche~\citep{lomonaco2021avalanche} and Continuum~\citep{douillard2021continuum} provide abstractions and baselines, while general domain-shift suites (e.g., WILDS~\citep{koh2021wilds}) and domain generalization toolkits (e.g., DomainBed~\citep{gulrajani2021in}) target classification or detection tasks. MMPose and Detectron2 offer strong pose tooling but assume static training. \textit{PoseAdapt complements these ecosystems by bringing pose-specific continual protocols, plug-in CL strategies, and unified evaluation under deployment-motivated constraints, enabling reproducible comparisons and practical adaptation pathways.}

\section{PoseAdapt Framework}
\label{sec:framework}

PoseAdapt is designed as a general-purpose continual adaptation layer on top of MMPose, enabling pretrained estimators to evolve across a sequence of experiences without requiring modifications to underlying backbones or datasets. It provides a unified mechanism for expressing how a model should initialize, update, and regularize itself as new experiences arrive. The goal is to decouple the mechanics of continual adaptation from the specifics of any particular backbone, dataset, or training regime.

\noindent \textbf{Problem Setting.}
We consider a stream of experiences
\(
\mathcal{E}_1,\;\mathcal{E}_2,\;\ldots,\; \mathcal{E}_T,
\)
where each experience $\mathcal{E}_i$ provides a dataset $\mathcal{D}_i$ drawn either from a new domain (e.g.\ lighting, modality) or an expanded keypoint set. A pretrained off-the-shelf pose estimator $\mathcal{M}_0$ serves as the starting point. At each stage, the framework produces an updated model
\(
\mathcal{M}_i = \Phi(\mathcal{M}_{i-1}, \mathcal{D}_i; \pi),
\)
where $\pi$ denotes the chosen continual learning strategy. This formulation explicitly separates three roles: the backbone architecture contained in $\mathcal{M}_{i-1}$, the new information encoded in $\mathcal{D}_i$, and the adaptation policy encoded in $\pi$. Each experience is decomposed into three stages: (1) \emph{Initialization}: prepare the structure and reference state required by the strategy; 
(2) \emph{Adaptation}: optimize parameters on $\mathcal{D}_i$ under constraints of $\pi$;  and
(3) \emph{Finalization}: compute strategy-specific statistics used by the next experience. \Cref{fig:poseadapt-framework} illustrates this cycle: a new experience triggers structural initialization and then constrained adaptation. The right side shows how head expansion preserves output-space compatibility during class-incremental updates.

\subsection{Initialization Phase}

Given $\mathcal{E}_i$, PoseAdapt prepares $\mathcal{M}_{i-1}$ for the new experience.

For fixed-architecture strategies (LwF, LFL, EWC), a frozen reference snapshot
\(
\tilde{\mathcal{M}}_{i-1}
\)
is created. This snapshot provides target features (LFL), target logits (LwF), or parameter anchors (EWC).

In class-incremental settings, the prediction head expands from $K_{i-1}$ to $K_i$:
\[
W_i = \big[\, W_{i-1} \;\; \Delta W_i \,\big],
\]
with $\Delta W_i$ initialized via configurable policy.

\subsection{Adaptation Phase}

Let $\theta_i$ denote the trainable parameters during experience $i$.  
The supervised loss on $\mathcal{D}_i$ is $\mathcal{L}_{\mathrm{kpt}}(\theta_i;\mathcal{D}_i)$.  
Continual learning adds a strategy-dependent regularizer:
\[
\mathcal{L}(\theta_i)
=
(1-\alpha)\,\mathcal{L}_{\mathrm{kpt}}
+
\alpha\,\mathcal{L}_{\mathrm{reg}}(\theta_i;\tilde{\mathcal{M}}_{i-1};\pi).
\]

\noindent \textbf{Less-Forgetful Learning (LFL).}
LFL constrains the feature extractor to preserve the geometry learned previously.  
Let $f_i(x)$ and $\tilde{f}_{i-1}(x)$ denote backbone feature maps from the current and teacher models.  
The penalty is the per-level MSE:
\[
\mathcal{L}_{\mathrm{reg}}^{\mathrm{LFL}}
=
\frac{1}{|\mathcal{B}|}
\sum_{x\in\mathcal{B}}
\sum_{\ell}
\big\| f_i^{(\ell)}(x) - \tilde{f}_{i-1}^{(\ell)}(x) \big\|_2^2.
\]

\noindent \textbf{Learning without Forgetting (LwF).}
LwF distills the teacher’s output behavior.  
Let $z_i(x)$ and $\tilde{z}_{i-1}(x)$ denote logits from the current and teacher models (aligned by an optional converter if keypoints change).  
PoseAdapt applies softened KL divergence with confidence masking:
\[
\mathcal{L}_{\mathrm{reg}}^{\mathrm{LwF}}
=
\tau^2\,
\frac{1}{|\mathcal{B}|}
\sum_{x\in\mathcal{B}}
\mathrm{KL}\!\left(
\sigma(z_i(x)/\tau)
\;\|\;
\sigma(\tilde{z}_{i-1}(x)/\tau)
\right).
\]

\noindent \textbf{EWC (online).}
EWC penalizes deviation from previous parameter values based on Fisher importances.  
Let $\theta_{i-1}$ be parameters after the last experience, and let $\hat{F}_{i-1,p}$ be the accumulated importance of parameter $p$.  
The penalty is
\[
\mathcal{L}_{\mathrm{reg}}^{\mathrm{EWC}}
=
\sum_{p}
\hat{F}_{i-1,p}\,
(\theta_{i,p}-\theta_{i-1,p})^2.
\]

\subsection{Finalization Phase}

After completing experience $i$, PoseAdapt computes and stores strategy-specific state: LFL and LwF update the teacher snapshot $\tilde{\mathcal{M}}_{i}$, whereas EWC computes fresh Fisher importances $F_{i,p}$ using a dedicated pass over $\mathcal{D}_i$:
\[
F_{i,p}
=
\frac{1}{|\mathcal{B}_i|}
\sum_{b\in\mathcal{B}_i}
\left(
\frac{\partial \mathcal{L}_{\mathrm{kpt}}(b)}{\partial \theta_{p}}
\right)^2,
\]
then update the accumulated values via
\(
\hat{F}_{i,p}=\lambda\,\hat{F}_{i-1,p}+F_{i,p}.
\)
For the initial model $\mathcal{M}_0$, importances are computed from the COCO validation split as we assume no access to pretraining data. These finalization steps produce the reference state required for experience $i+1$ and make strategy behavior explicit and reproducible.
\section{Benchmark and Experiments}
\label{sec:benchmark}

PoseAdapt defines a controlled evaluation setting for continual adaptation in 2D human pose estimation with two complementary tracks: \emph{domain‑incremental} and \emph{class‑incremental}. The goal is to isolate the behavior of continual learning (CL) strategies under deployment‑motivated constraints while holding capacity and data access fixed.

\subsection{Experimental Setup}
\label{sec:protocols}

\noindent \textbf{Methods.}
We evaluate \emph{naïve fine-tuning} (FT), \emph{Elastic Weight Consolidation} (EWC)~\cite{kirkpatrick2017overcoming}, \emph{Less-Forgetful Learning} (LFL)~\cite{jung2016less}, and \emph{Learning without Forgetting} (LwF)~\cite{li2017learning}.

\begin{figure}[th]
    \centering
    \includegraphics[width=0.495\linewidth]{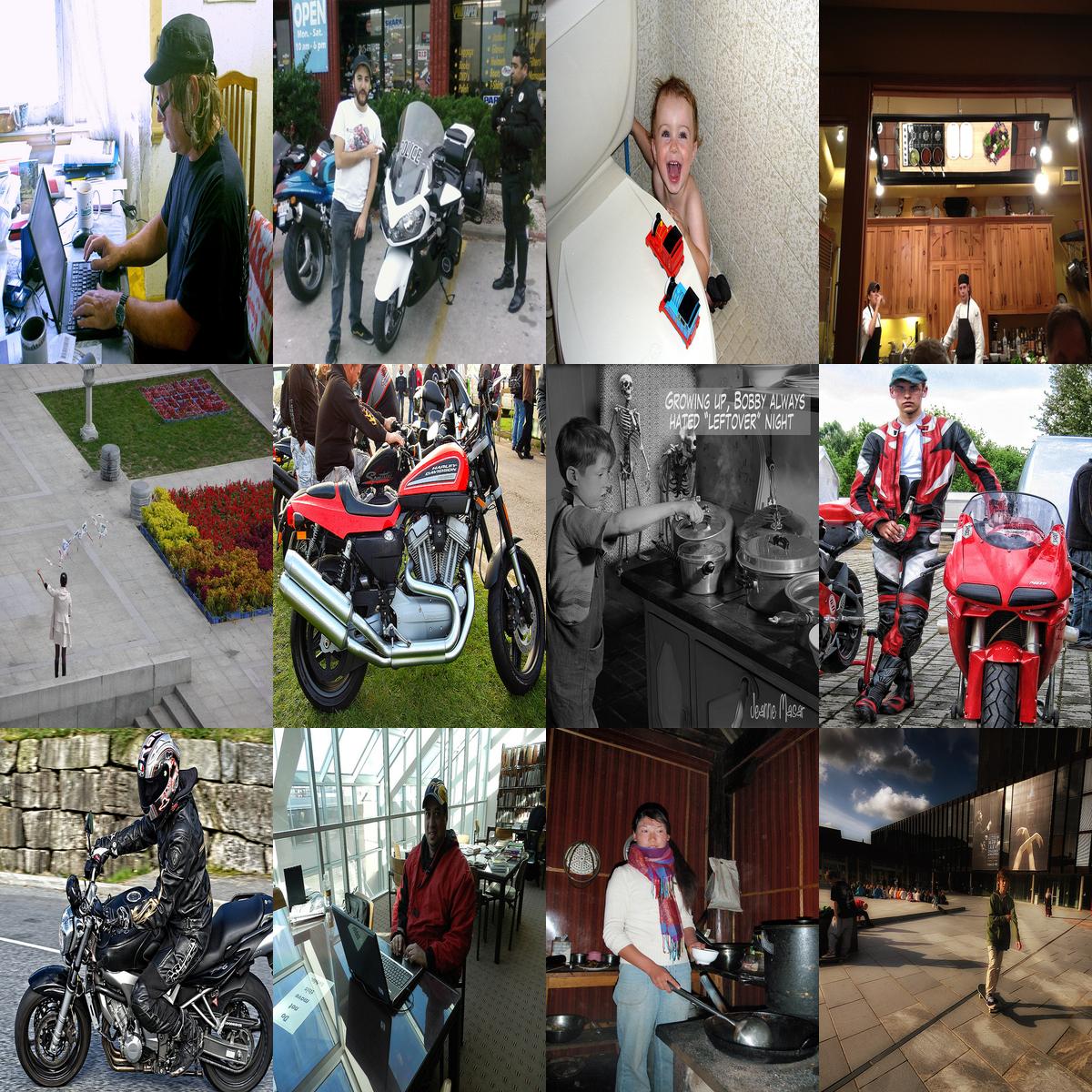}
    \includegraphics[width=0.495\linewidth]{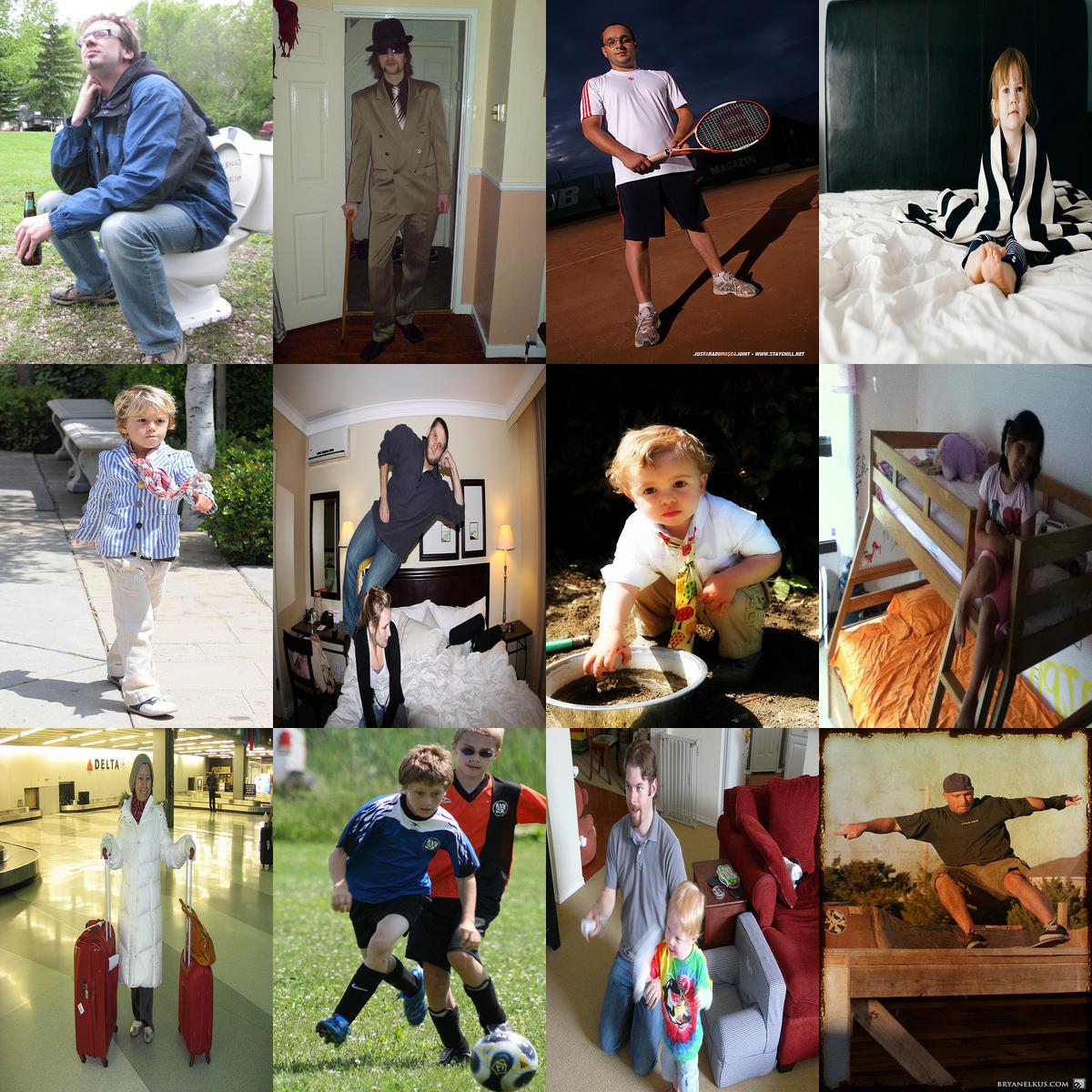}
    \caption{\textbf{Reference domain.} Examples from the COCO validation set representing the well-lit RGB baseline used for the initial experience in all benchmarks.}
    \label{fig:benchmark-data-reference}
\end{figure}

\noindent \textbf{Reference model and data.} All methods adapt the same off-the-shelf top-down pose estimator. To isolate keypoint regression from person detection, \emph{ground‑truth detection boxes} are used. The model (RTMPose‑t~\cite{jiang2023rtmpose}, $\sim$3M parameters) is pretrained on the COCO~\cite{lin2014microsoft} and AIC datasets to predict \emph{17 keypoints} with \emph{70.06 AP} on the COCO validation set\footnote{\url{https://download.openmmlab.com/mmpose/v1/projects/rtmposev1/rtmpose-tiny_simcc-aic-coco_pt-aic-coco_420e-256x192-cfc8f33d_20230126.pth}}. We denote this as the reference domain (\cref{fig:benchmark-data-reference}) in the first experience which the model must not forget as it adapts to new experiences with various domain shifts.

\noindent \textbf{Metrics.}
The mean average precision (AP) at the end of training all experiences is reported. In addition, following standard continual learning evaluation~\cite{lopez2017gradient}, we report retention accuracy (RA) and average forgetting (AF). Let \(a_{i,j}\) denote the AP on experience \(j\) after training on experience \(i\). For a sequence of \(T\) experiences:
\begin{align}
\mathrm{RA} &= \tfrac{1}{T}\sum_{t=1}^{T} a_{T,t}, \\
\mathrm{AF} &= \tfrac{1}{T-1}\sum_{t=1}^{T-1} \left(a_{t,t} - a_{T,t}\right).
\end{align}
RA measures how well the final model performs across all experiences, while AF quantifies the average drop in performance on earlier experiences after completing. When interpreted together, RA reflects overall stability, and AF indicates how much this stability depends on preserving early experiences (i.e., backward transfer). High RA combined with low AF indicates effective continual learning.

\noindent \textbf{Hyperparameters and constraints.}
Adaptation uses AdamW~\cite{loshchilov2017decoupled} with a base learning rate 0.004, no weight decay, random seed 21, and a small linear warm-up. All images are normalized with $\mu=[123.675, 116.28, 103.53], \sigma=[58.395, 57.12, 57.375]$ and augmented using random horizontal flips and half-body occlusions. Each experience is allocated at most 1k labeled images and 10 epochs. Past data are never stored or revisited. The only state retained across experiences is the most recent network; distillation‑based methods additionally keep a teacher snapshot from the previous experience. For EWC, the diagonal Fisher and parameter means are computed at the end of each experience on its training stream and stored for the next step.

\subsection{Domain-Incremental Track}
\label{sec:domain-incremental}

The domain-incremental track evaluates the ability to adapt across progressively more challenging shifts while retaining performance on previously seen domains. All domain shifts are generated post‑hoc from the reference data to ensure reproducibility under shared identities and annotations.

\subsubsection{Scene Density}

\begin{figure}[ht]
    \centering
    \setlength{\tabcolsep}{0.5pt}
    \scriptsize
    \begin{tabularx}{\linewidth}{YYY}
        \includegraphics[width=\linewidth]{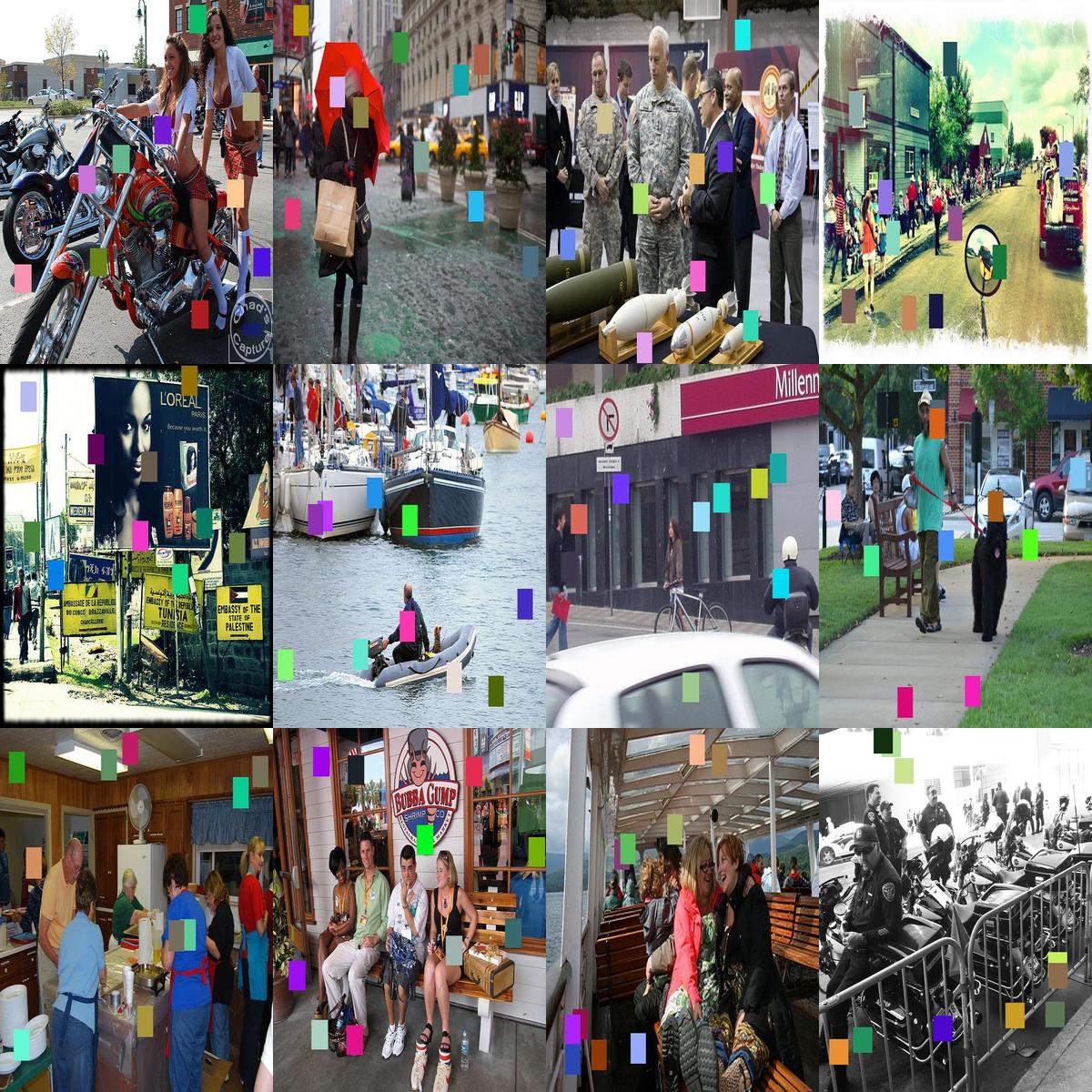}
        &
        \includegraphics[width=\linewidth]{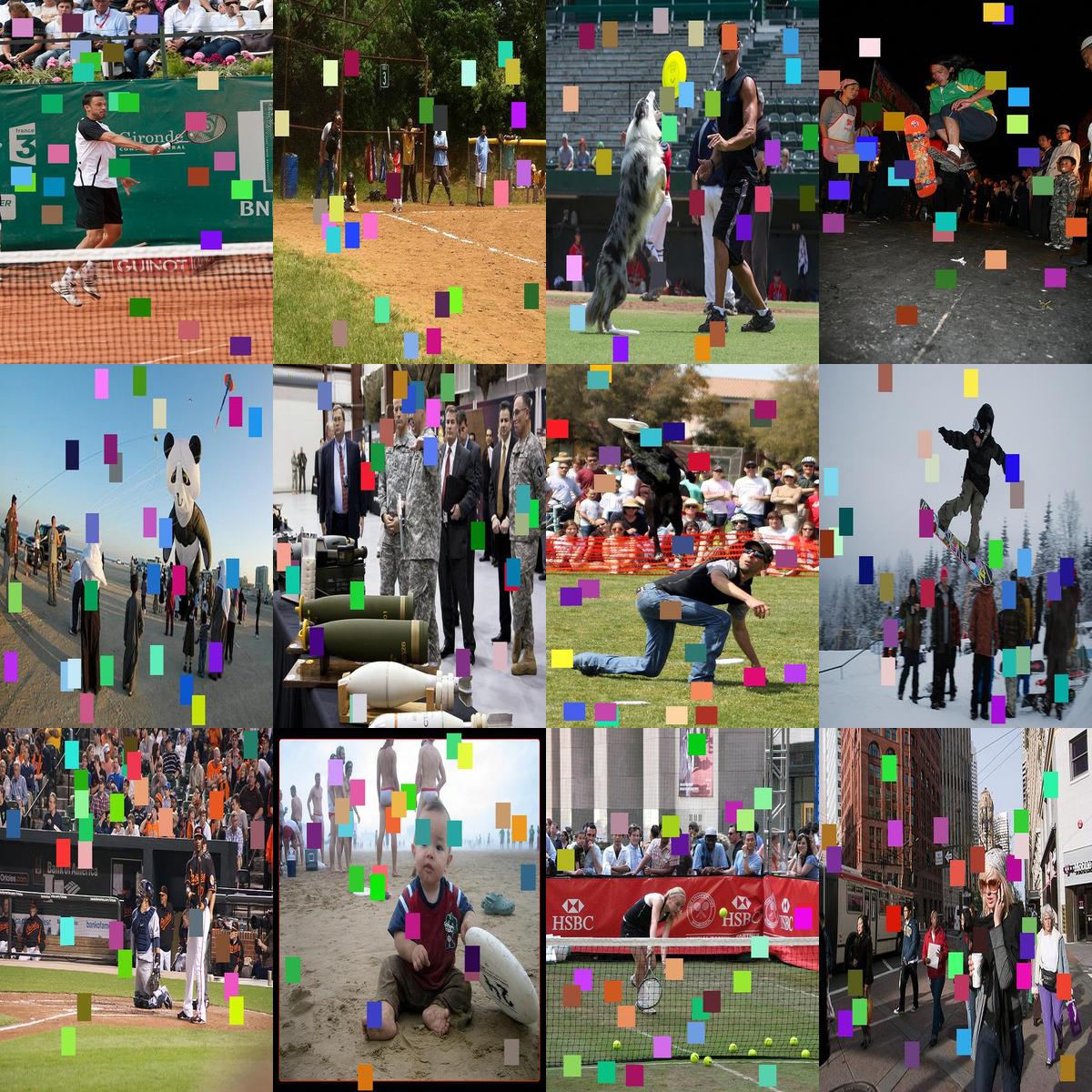}
        &
        \includegraphics[width=\linewidth]{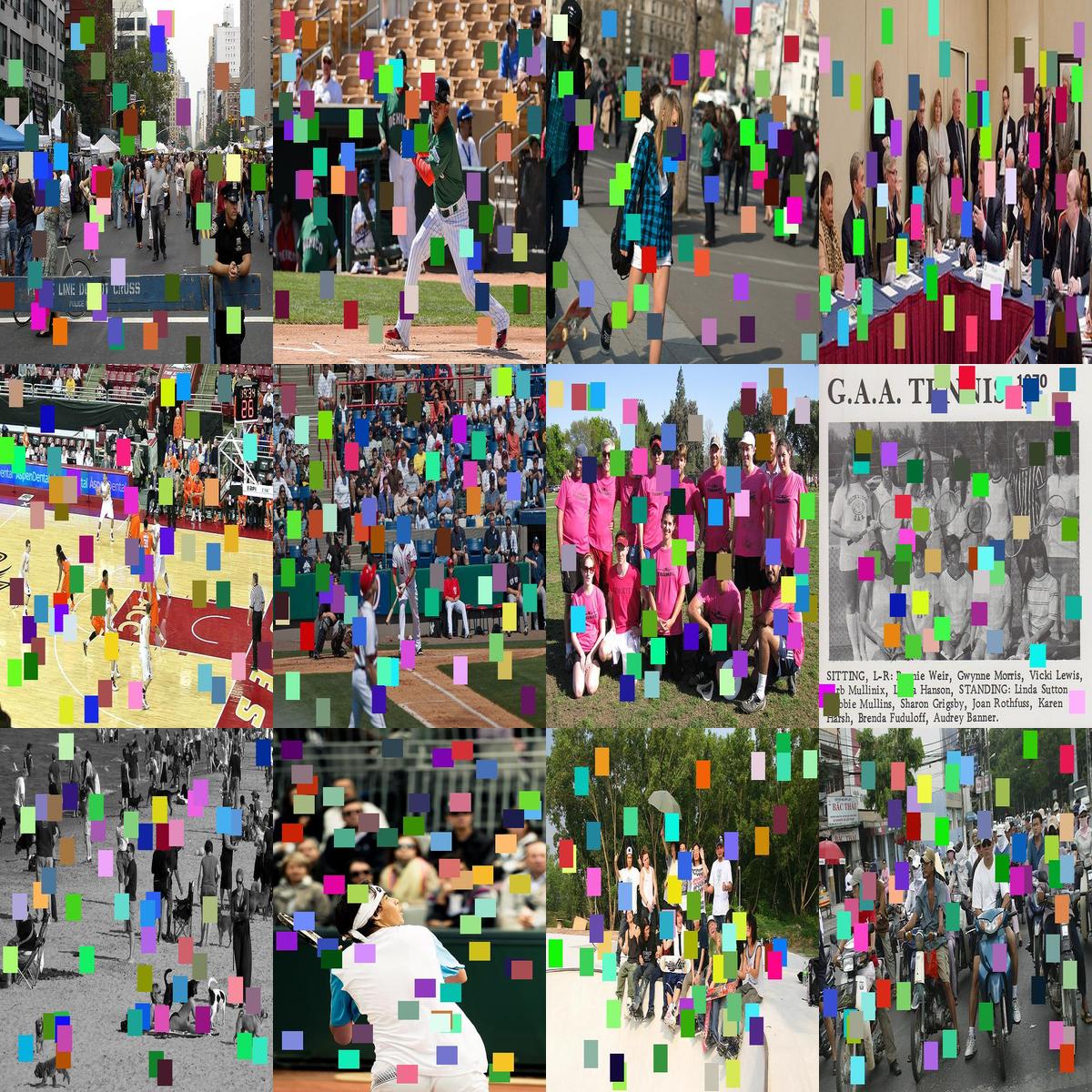}
        \\
        \textbf{Light} (\textcolor{red}{O5}) &
        \textbf{Medium} (\textcolor{red}{O10}) &
        \textbf{High} (\textcolor{red}{O20}) \\
    \end{tabularx}
    \caption{\textbf{Occlusion data.} Increasing scene density combined with synthetic cutout occlusion produces the three density-shift experiences: O5 (light), O10 (medium), and O20 (heavy).}
    \label{fig:benchmark-data-di-occlusion}
\end{figure}

\noindent \textbf{Shift generation.}
Scene density shifts are created in two stages: image selection followed by synthetic occlusion. For each difficulty level, we first filter COCO images by the number of annotated people: 3–6, 7–10, and $\ge$11 individuals respectively (with minimum keypoint counts of 10/1/1). This yields increasingly crowded scenes while preserving original COCO annotations. Next, we apply fixed-budget cutout occlusion to each image: square blocks of random color are stamped at random locations such that approximately 5\%, 10\%, or 20\% of the image area is occluded, using 10, 25, or 50 blocks per image. Together, these choices produce three reproducible density experiences—\emph{O5}, \emph{O10}, \emph{O20}—with increasing scene density as shown in~\cref{fig:benchmark-data-di-occlusion}.

\noindent \textbf{Results.}
Across O5, O10, and O20, all methods exhibit moderate forgetting, with stability decreasing as crowding and occlusion intensify.
In single-step adaptation, LwF is most stable for light occlusion (AF = 13.16) and maintains the highest RA there (56.88). Under medium and heavy occlusion, LFL becomes the most reliable, giving the lowest forgetting (AF = 13.58/14.23) and highest RA (50.41/52.82).
In the sequential setting, accuracy drops compared to the pretrained reference (RA = 59.06), but LFL is consistently the strongest regularizer with RA = 51.02, while LwF achieves the lowest forgetting (AF = 5.97) but slightly lower RA.

\begin{table*}[ht]
\begin{subtable}{\linewidth}
    \adjustbox{max width=\linewidth}{
    \begin{tabular}{lccggccggccggccccgg}
    \toprule
    \rowcolor{gray!20}
    \textbf{DENSITY} &
    \multicolumn{4}{c}{\textbf{Light}} &
    \multicolumn{4}{c}{\textbf{Medium}} &
    \multicolumn{4}{c}{\textbf{Heavy}} &
    \multicolumn{6}{c}{\textbf{Sequential}} \\
    \cmidrule(lr){1-1} \cmidrule(lr){2-5} \cmidrule(lr){6-9} \cmidrule(lr){10-13} \cmidrule(lr){14-19}
    Method & 
    \textcolor{blue}{N} & \textcolor{red}{O5} & AF $\downarrow$ & RA $\uparrow$ &
    \textcolor{blue}{N} & \textcolor{red}{O10} & AF $\downarrow$ & RA $\uparrow$ &
    \textcolor{blue}{N} & \textcolor{red}{O20} & AF $\downarrow$ & RA $\uparrow$ & \textcolor{blue}{N} & \textcolor{red}{O5} & \textcolor{red}{O10} & \textcolor{red}{O20} & AF $\downarrow$ & RA $\uparrow$ \\
    \cmidrule(lr){1-1} \cmidrule(lr){2-5} \cmidrule(lr){6-9} \cmidrule(lr){10-13} \cmidrule(lr){14-19}
    FT         & 53.92 & 54.18 & 16.14 & 54.05
               & 52.92 & 42.41 & 17.15 & 47.66
               & 51.90 & 48.70 & 18.16 & 50.30
               & 52.35 & 53.39 & 42.54 & 49.69 & 6.16 & 49.49 \\
    \cmidrule(lr){2-5} \cmidrule(lr){6-9} \cmidrule(lr){10-13} \cmidrule(lr){14-19}
    EWC        & 53.04 & 52.25 & 17.02 & 52.65
               & 53.23 & 40.94 & 16.83 & 47.09
               & 51.78 & 47.42 & 18.29 & 49.60
               & 51.81 & 53.15 & 41.89 & 48.86 & 6.13 & 48.93 \\
    LFL        & 56.78 & 56.78 & 13.29 & 56.78
               & \textbf{56.49} & \textbf{44.34} & \textbf{13.58} & \textbf{50.41}
               & \textbf{55.83} & \textbf{49.81} & \textbf{14.23} & \textbf{52.82}
               & \textbf{54.27} & 54.96 & \textbf{44.29} & \textbf{50.57} & 6.31 & \textbf{51.02} \\
    LwF        & \textbf{56.91} & \textbf{56.86} & \textbf{13.16} & \textbf{56.88}
               & 54.97 & 42.92 & 15.10 & 48.94
               & 54.81 & 49.61 & 15.25 & 52.21
               & 53.92 & \textbf{55.45} & 43.33 & 49.80 & \textbf{5.97} & 50.62 \\
    \end{tabular}
    }
\end{subtable}
\begin{subtable}{\linewidth}
    \adjustbox{max width=\linewidth}{
    \begin{tabular}{lccggccggccggccccgg}
    \midrule
    \rowcolor{gray!20}
    \textbf{LIGHTING} &
    \multicolumn{4}{c}{\textbf{Low}}
    &
    \multicolumn{4}{c}{\textbf{Very Low}} &
    \multicolumn{4}{c}{\textbf{Extremely Low}} &
    \multicolumn{6}{c}{\textbf{Sequential}} \\
    \cmidrule(lr){1-1} \cmidrule(lr){2-5} \cmidrule(lr){6-9} \cmidrule(lr){10-13} \cmidrule(lr){14-19}
    Method & \textcolor{blue}{WL} & \textcolor{red}{LL} & AF $\downarrow$ & RA $\uparrow$ & \textcolor{blue}{WL} & \textcolor{red}{VLL} & AF $\downarrow$ & RA $\uparrow$ & \textcolor{blue}{WL} & \textcolor{red}{ELL} & AF $\downarrow$ & RA $\uparrow$ & \textcolor{blue}{WL} & \textcolor{red}{LL} & \textcolor{red}{VLL} & \textcolor{red}{ELL} & AF $\downarrow$ & RA $\uparrow$ \\
    \cmidrule(lr){1-1} \cmidrule(lr){2-5} \cmidrule(lr){6-9} \cmidrule(lr){10-13} \cmidrule(lr){14-19}
    FT         & 53.83 & 51.45 & 16.24 & 52.64
               & 48.51 & 46.74 & 21.56 & 47.62
               & 27.55 & 40.07 & 42.52 & 33.81
               & 36.99 & 40.29 & 43.10 & 39.09 & 15.86 & 39.87 \\
    \cmidrule(lr){2-5} \cmidrule(lr){6-9}  \cmidrule(lr){10-13} \cmidrule(lr){14-19}
    EWC        & 52.63 & 50.15 & 17.43 & 51.39
               & 45.92 & 43.21 & 24.15 & 44.56
               & 20.74 & 40.64 & 49.32 & 30.69
               & 28.65 & 36.84 & 41.92 & 39.45 & 19.51 & 36.72 \\
    LFL        & \textbf{59.16} & \textbf{56.47} & \textbf{10.91} & \textbf{57.81}
               & \textbf{54.60} & \textbf{51.04} & \textbf{15.47} & \textbf{52.82}
               & \textbf{39.93} & 43.22 & \textbf{30.14} & \textbf{41.57}
               & \textbf{37.42} & \textbf{43.90} & \textbf{46.54} & \textbf{40.73} & \textbf{15.58} & \textbf{42.15} \\
    LwF        & 58.37 & 55.74 & 11.69 & 57.06
               & 52.07 & 49.00 & 18.00 & 50.53
               & 38.87 & \textbf{43.53} & 31.20 & 41.20
               & 26.17 & 34.01 & 39.71 & 37.51 & 24.01 & 34.35 \\
    \end{tabular}
    }
\end{subtable}
\begin{subtable}{\linewidth}
    \adjustbox{max width=\linewidth}{
    \scriptsize
    \begin{tabular}{lccggccggcccgg}
    \midrule
    \rowcolor{gray!20}
    \textbf{MODALITY} &
    \multicolumn{4}{c}{\textbf{Grayscale}}
    &
    \multicolumn{4}{c}{\textbf{Depth Image}} &
    \multicolumn{5}{c}{\textbf{Sequential}} \\
    \cmidrule(lr){1-1} \cmidrule(lr){2-5} \cmidrule(lr){6-9}  \cmidrule(lr){10-14}
    Method & \textcolor{blue}{RGB} & \textcolor{red}{Gray} & AF $\downarrow$ & RA $\uparrow$ & \textcolor{blue}{RGB} & \textcolor{red}{Depth} & AF $\downarrow$ & RA $\uparrow$ & \textcolor{blue}{RGB} & \textcolor{red}{Gray} & \textcolor{red}{Depth} & AF $\downarrow$ & RA $\uparrow$ \\
    \cmidrule(lr){1-1} \cmidrule(lr){2-5} \cmidrule(lr){6-9}  \cmidrule(lr){10-14}
    FT         & 48.25 & 50.70 & 21.82 & 49.47
               &  6.66 & 35.43 & 63.40 & 21.05
               &  5.45 &  4.90 & 35.42 & 55.17 & 15.26 \\
    \cmidrule(lr){2-5} \cmidrule(lr){6-9}  \cmidrule(lr){10-14}
    EWC        & 47.87 & 48.92 & 22.19 & 48.39
               &  \textbf{8.68} & 36.17 & \textbf{61.39} & \textbf{22.42}
               & \textbf{13.92} & \textbf{12.60} & 35.19 & \textbf{47.52} & \textbf{20.57} \\
    LFL        & \textbf{48.80} & \textbf{53.46} & \textbf{21.26} & \textbf{51.13}
               &  6.72 & 32.74 & 63.34 & 19.73
               & 12.48 &  9.53 & 35.13 & 50.33 & 19.05 \\
    LwF        & 46.68 & 50.55 & 23.39 & 48.61
               &  5.41 & \textbf{37.49} & 64.66 & 21.45
               & 13.23 & 10.29 & \textbf{37.36} & 50.14 & 20.30 \\
    \bottomrule
    \end{tabular}
    }
\end{subtable}
\caption{\textbf{PoseAdapt Domain-Incremental Results.} Accuracy (AP) of continual learning strategies across three domain-incremental
benchmarks: Scene Density (increasing crowding/occlusion), \emph{Lighting} (progressively darker conditions), and \emph{Modality} (RGB → grayscale,
depth, silhouette). \textcolor{blue}{Blue} entries denote the reference/normal domain, while \textcolor{red}{Red} entries denote the target/shifted domains requiring adaptation. $RA$ indicates the mean AP across listed domains at the end of training. FT is the naive finetuning. Because
the benchmark enforces a fixed lightweight backbone and a strict adaptation budget, fine-tuning (FT) can underperform a frozen pretrained model, even on the latest domain. This is expected and highlights the difficulty of sustainable adaptation under constrained resources. Regularization-based methods are compared under single-step and sequential (online) protocols.}
\label{tab:results-di}
\end{table*}

\subsubsection{Lighting}

\begin{figure}[ht]
    \centering
    \setlength{\tabcolsep}{0.5pt}
    \scriptsize
    \begin{tabularx}{\linewidth}{YYY}
        \includegraphics[width=\linewidth]{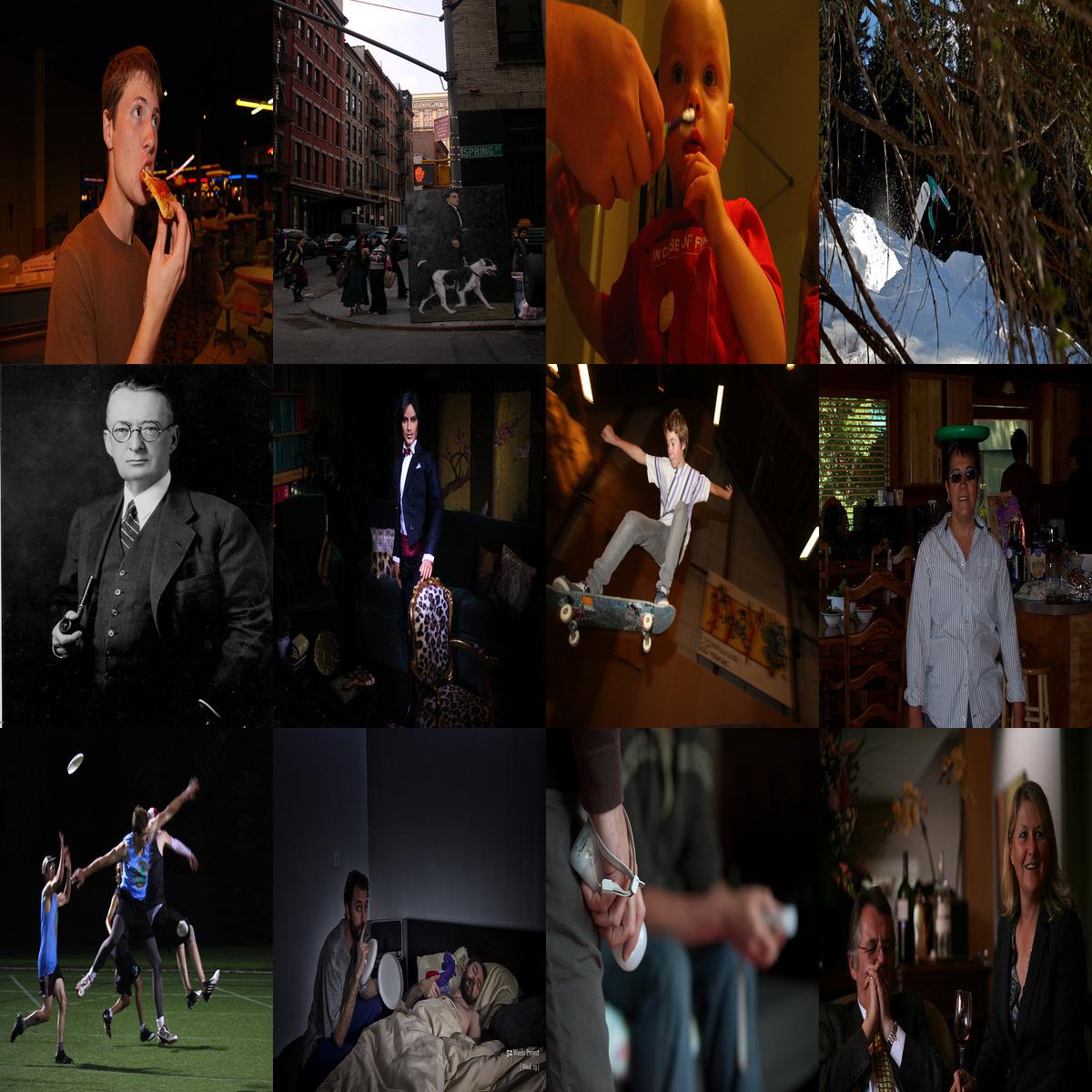}
        &
        \includegraphics[width=\linewidth]{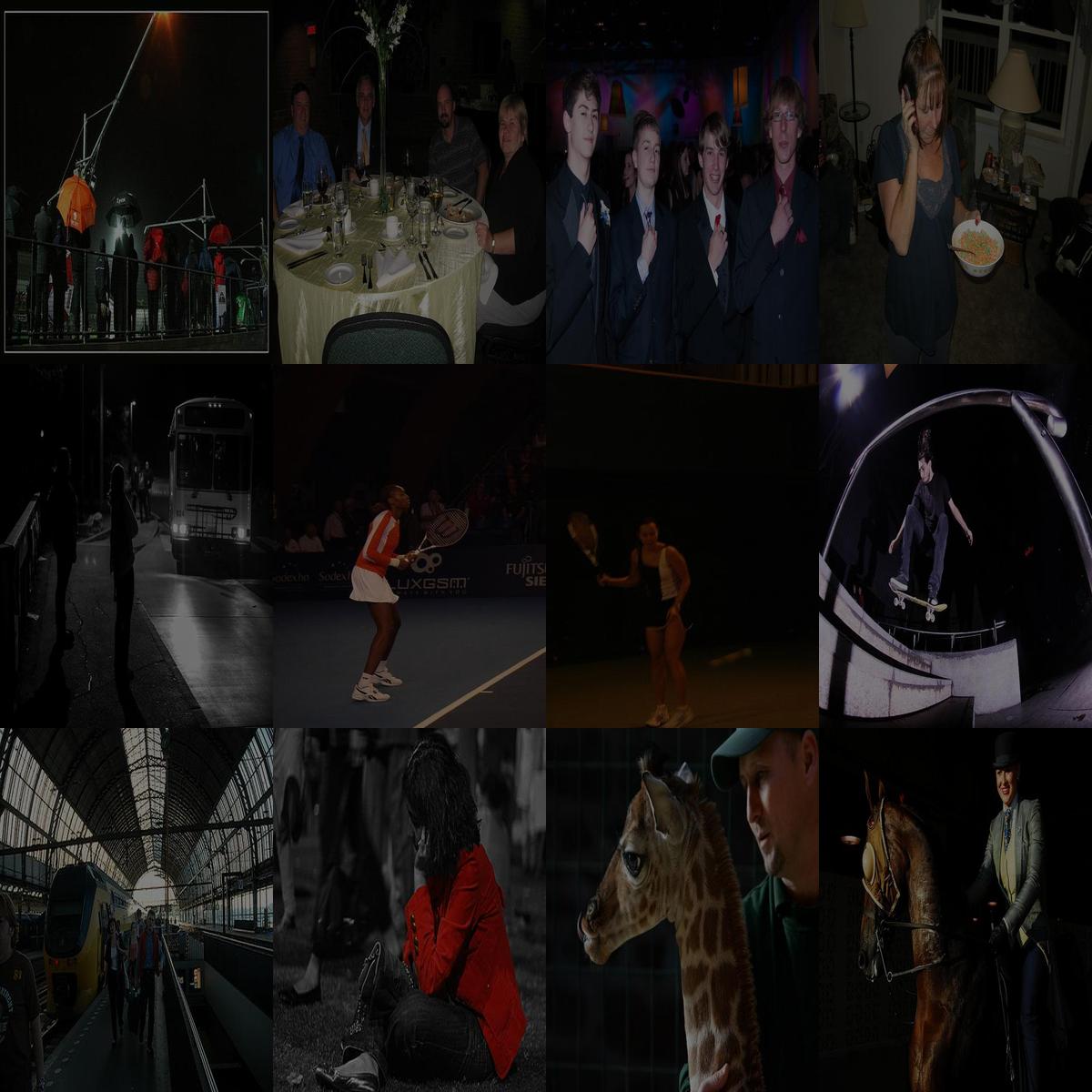}
        &
        \includegraphics[width=\linewidth]{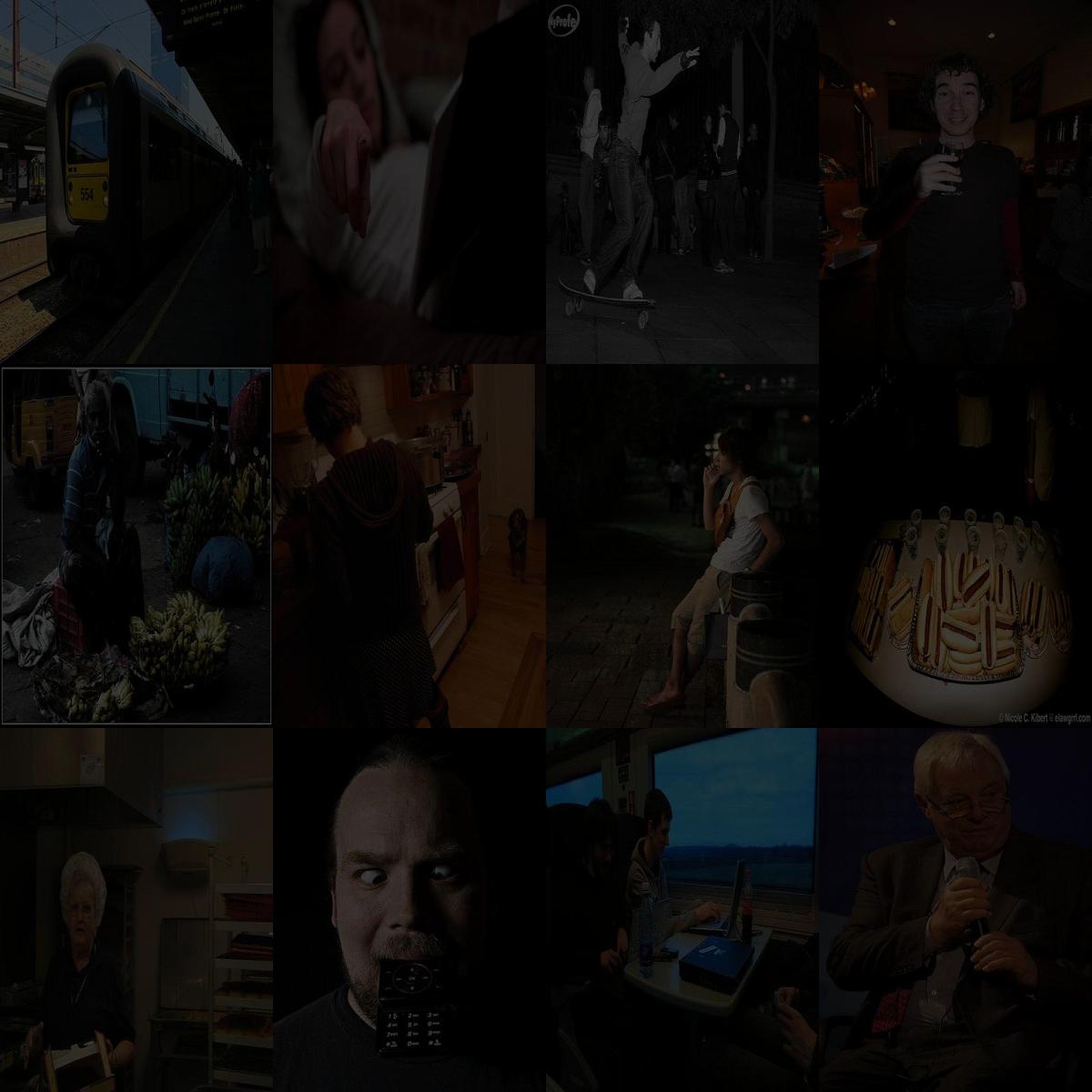}
        \\
        \textbf{Low} (\textcolor{red}{LL}) &
        \textbf{Very Low} (\textcolor{red}{VLL}) &
        \textbf{Extremely Low} (\textcolor{red}{ELL}) \\
    \end{tabularx}
    \caption{\textbf{Lighting data.} This consists of three levels of decreasing illumination—LL, VLL, and ELL—obtained through brightness-based selection and controlled photometric darkening.}
    \label{fig:benchmark-data-di-lighting}
\end{figure}

\noindent \textbf{Shift generation.}
Low-light images (\emph{LL}) are obtained by scoring every COCO image (with 5+ annotated keypoints) using a brightness metric that averages (i) the global grayscale brightness of the full image and (ii) the grayscale brightness inside all person bounding boxes. This joint score emphasizes scenes where both the environment and the subjects are dark. The 1,000 images with the lowest brightness scores form the LL split. Progressively darker variants—\emph{VLL} and \emph{ELL}—are then synthesized from LL by applying controlled photometric degradations, primarily brightness reduction and contrast modification with very mild noise/blur, yielding increasing darkness levels shown in~\cref{fig:benchmark-data-di-lighting}.

\noindent \textbf{Results.}
Lighting shifts are considerably harder than density.
In single-step adaptation, LFL again provides the best stability at every darkness level, with AF = 10.91/15.47/30.14 for LL/VLL/ELL. FT adapts aggressively to the new domain but heavily sacrifices the well-lit reference (e.g., WL→27.55 under ELL). LwF tracks LFL under LL but degrades more sharply at VLL and ELL.
Sequentially, forgetting compounds strongly across LL→VLL→ELL. LFL remains the most robust method (RA = 42.15), with FT next (39.87), while EWC and LwF trail (36.72 and 34.35). The pattern reflects an increasingly nonlinear stability–plasticity trade-off as illumination diminishes.

\subsubsection{Modality}

\begin{figure}[ht]
    \centering
    \setlength{\tabcolsep}{0.5pt}
    \scriptsize
    \begin{tabularx}{\linewidth}{YY}
        \includegraphics[width=\linewidth]{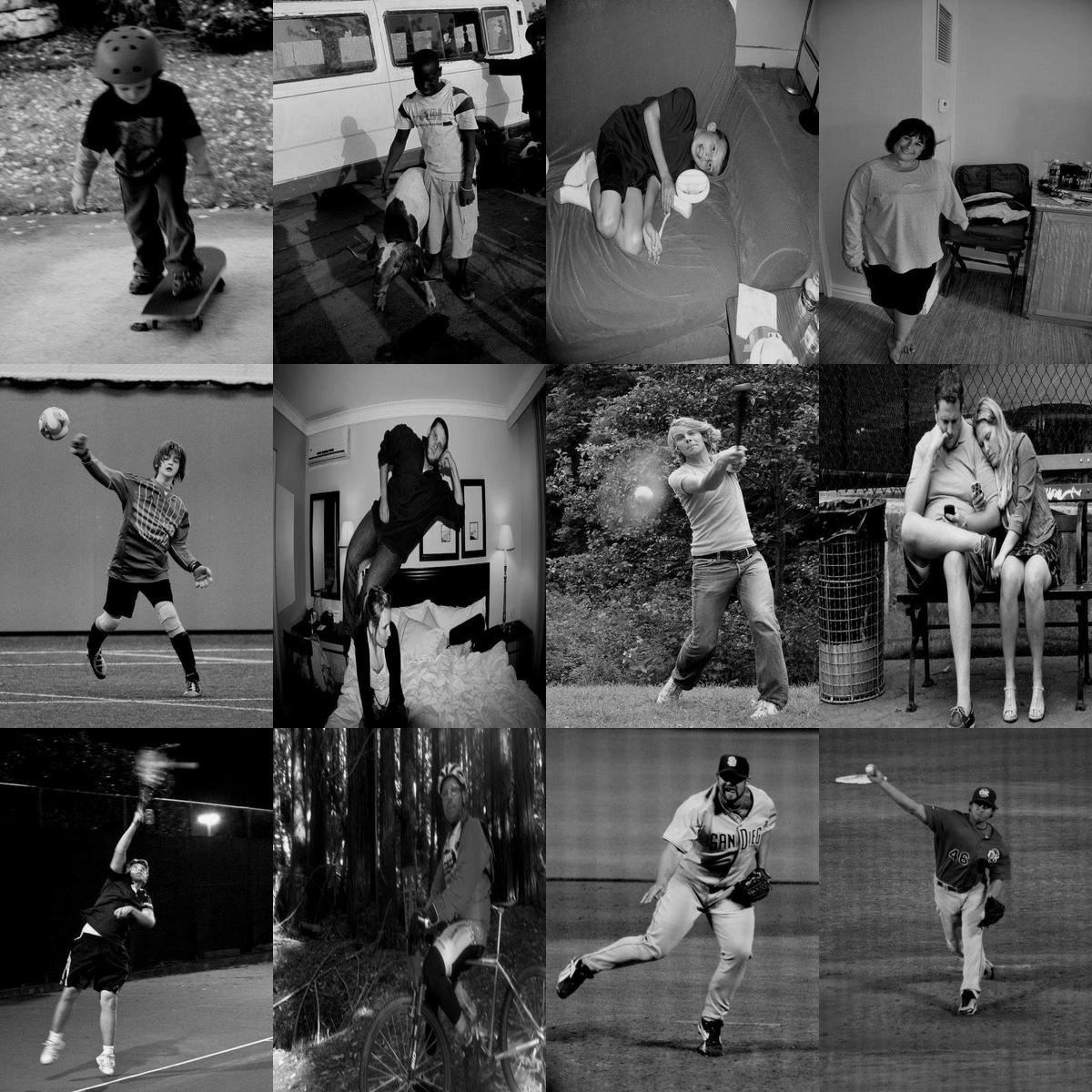}
        &
        \includegraphics[width=\linewidth]{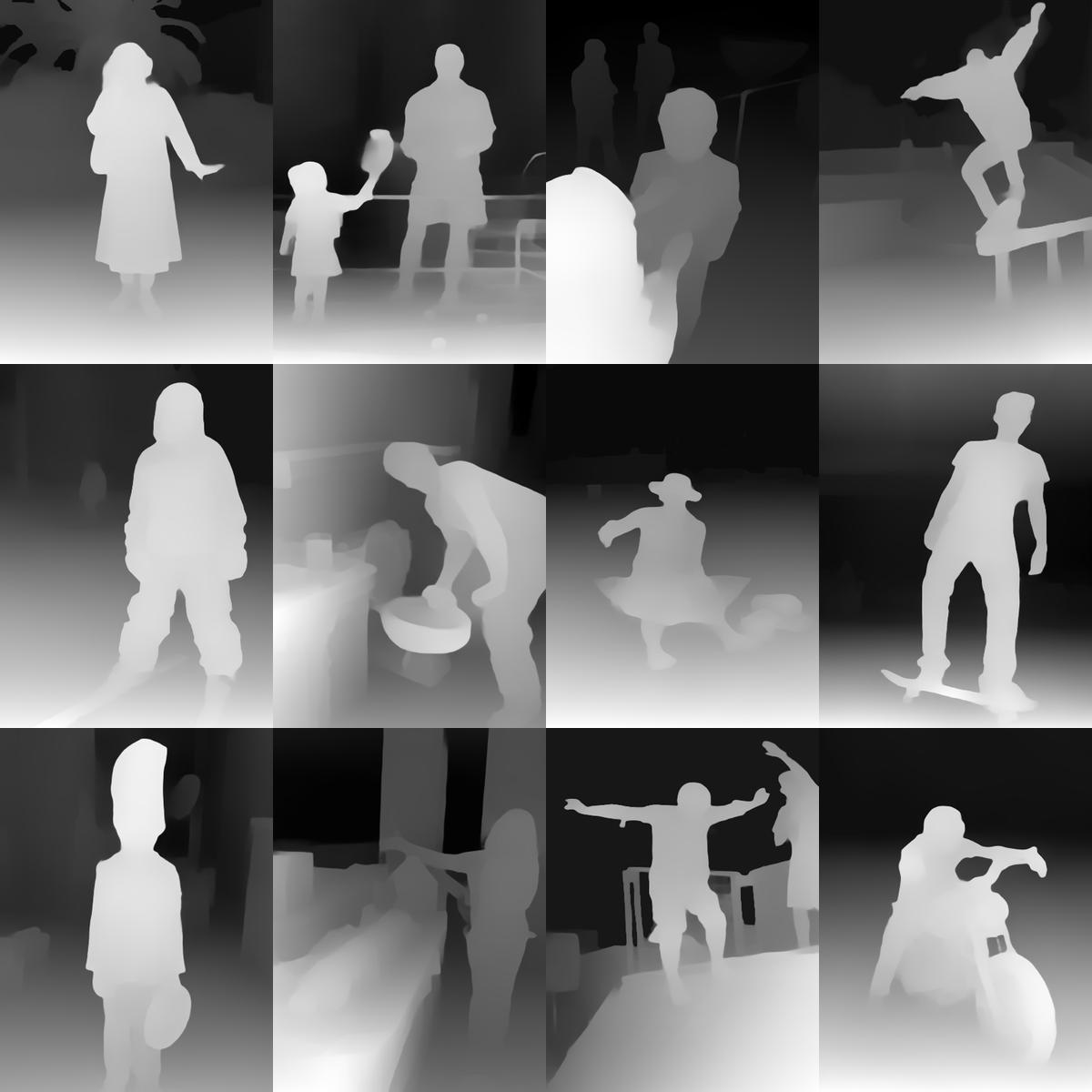}
        \\
        \textbf{Grayscale} (\textcolor{red}{Gray}) &
        \textbf{Depth Maps} (\textcolor{red}{Depth}) \\
    \end{tabularx}
    \caption{\textbf{Modality data.} The modality benchmark includes Gray (desaturated and perturbed grayscale) and Depth (MiDaS-derived relative-depth maps scaled to [0, 1]), both tiled to three channels.}
    \label{fig:benchmark-data-di-mode}
\end{figure}

\noindent \textbf{Shift generation.}
Modality shifts comprise two experiences: \emph{Gray} and \emph{Depth}. Grayscale images are produced by fully desaturating each RGB frame and then applying a sequence of mild photometric perturbations including some Gaussian noise, increased contrast, and a small brightness correction. Depth images are generated by running MiDaS (DPT-Large)~\cite{ranftl2020midas} per frame, resizing predictions back to the original resolution, and min–max normalizing the relative-depth map to [0,1]. Both outputs, illustrated in~\cref{fig:benchmark-data-di-mode}, are tiled to three channels for backbone compatibility.

\begin{figure*}
    \begin{subfigure}{0.55\linewidth}
        \centering
        \includegraphics[width=\linewidth]{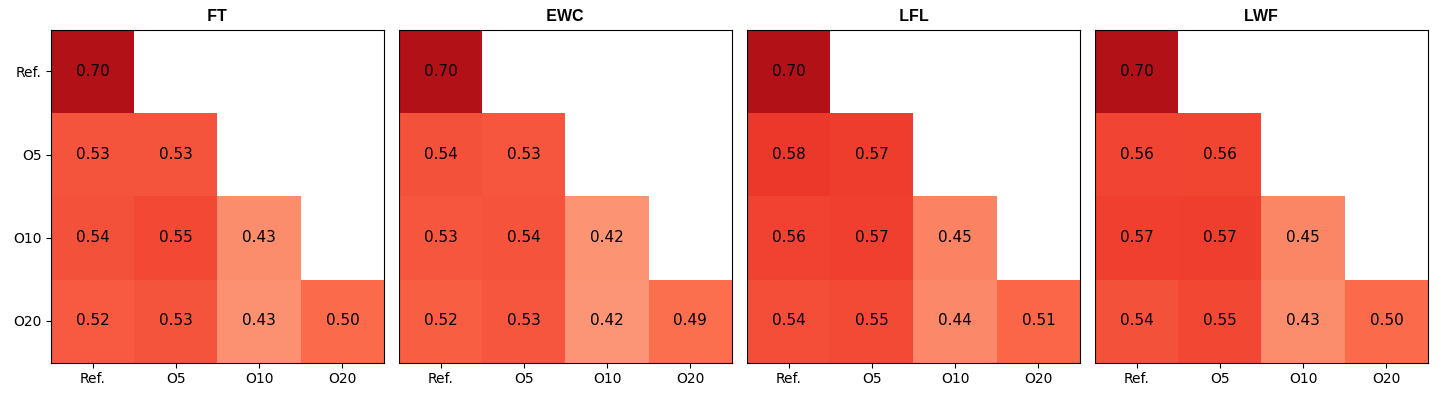}
        \caption{\emph{PoseAdapt-Density}}
        \label{fig:heatmap-density}
        \includegraphics[width=\linewidth]{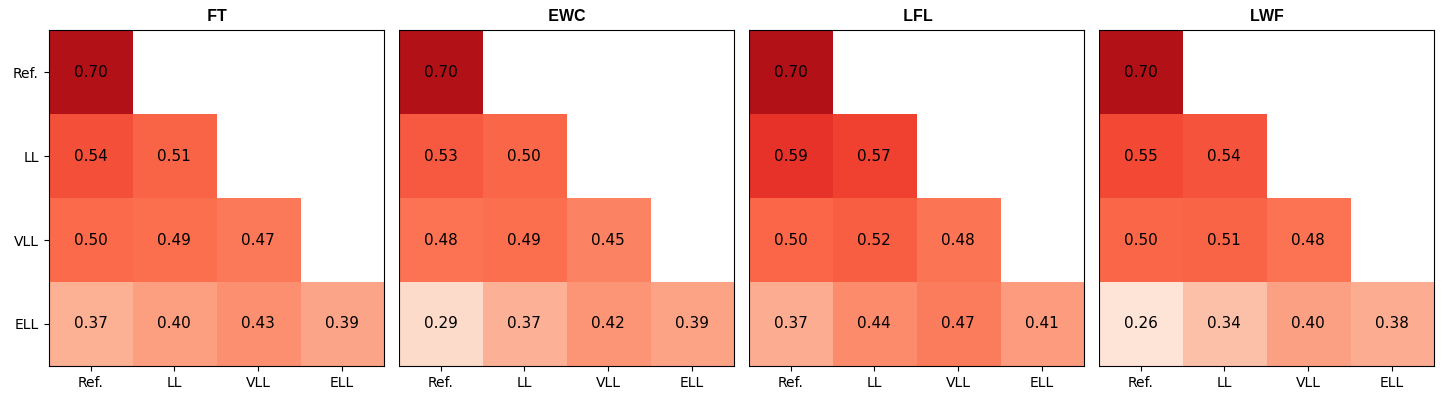}
        \caption{\emph{PoseAdapt-Lighting}}
        \label{fig:heatmap-lighting}
    \end{subfigure}
    \hfill
    \begin{subfigure}{0.445\linewidth}
    \centering
    \includegraphics[width=\linewidth]{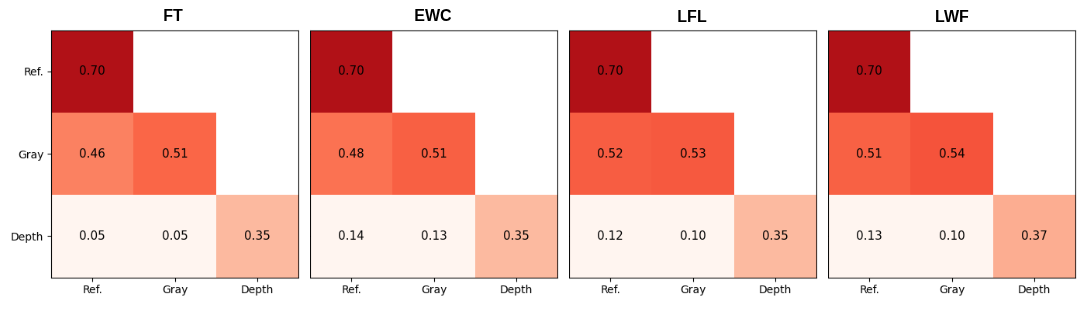}
    \caption{PoseAdapt-Modality}
    \label{fig:heatmap-mode}
    \plotExperiment{
        (0, 70.06)(2,60.38)(4,56.57)(6,54.35)(8,55.73)(10,54.08)(12,49.40)(14,50.54)(16,45.78)(18,49.54)(20,49.89)(22,40.59)(24,39.09)(26,38.81)(28,29.16)(30,36.99)
    }{
        (0,48)(2,56.83)(4,52.63)(6,50.72)(8,52.48)(10,50.77)(12,51.75)(14,51.39)(16,48.49)(18,49.50)(20,49.21)(22,45.06)(24,42.71)(26,41.80)(28,36.83)(30,40.29)
    }{
        (10,45)(12,48.56)(14,47.89)(16,44.69)(18,45.66)(20,47.14)(22,46.45)(24,44.17)(26,42.57)(28,41.10)(30,43.10)
    }{
        (20,36)(22,41.81)(24,39.91)(26,37.50)(28,37.20)(30,39.09)
    }{}
    \caption{Sequential adaptation curves for the lighting benchmark.}
    \label{fig:finetuning-trend}
\end{subfigure}
\caption{\textbf{Experience–experience performance matrices for the three domain-incremental benchmarks.}
Heatmaps (a–c) visualize forgetting and retention across Density, Lighting, and Modality. 
Rows denote the training experience, columns the evaluation experience, and diagonal cells show immediate post-training AP. 
Off-diagonal decay illustrates how quickly methods forget earlier domains as adaptation progresses. 
Density produces only mild degradation, Lighting shows progressively stronger drift, and Modality exhibits the steepest collapse—especially after Depth. 
(d) plots the sequential WL$\rightarrow$LL$\rightarrow$VLL$\rightarrow$ELL trajectory, highlighting the compounding loss on well-lit images as illumination decreases.}

\end{figure*}

\noindent \textbf{Results.}
Modality shifts are the most severe.
For grayscale, stability is modest: FT and EWC forget RGB strongly (AF $\approx$ 22), whereas LFL is most stable (AF = 21.26) and gives the highest Gray-domain AP (53.46).
Depth presents an extreme shift: LwF yields the best Depth AP (37.49), while EWC retains RGB best (AF = 61.39, still catastrophic). Sequentially, no method maintains usable RGB performance when moving through Gray→Depth: RA collapses to 15–20 across all methods (best: EWC at 20.57). This confirms the large appearance and geometric mismatch between RGB images and monocular depth maps.

\subsection{Class‑Incremental Track}

PoseAdapt supports skeleton growth scenarios analogous to class-incremental continual learning. New keypoints are introduced over time whereas labels for some past keypoints might not be available anymore. The model is expected to remember all seen keypoints while incorporating new ones to predict an expanded skeleton at each experience.

For standard evaluations of pure class-incremental continual pose estimation, we propose the PoseAdapt-BodyParts benchmark. In this benchmark, the network must progressively learn body, feet, hands, face, and spine keypoints. Body annotations with 17 keypoints from the COCO dataset~\cite{lin2014microsoft}, feet, face, and hand annotations with 6, 68, and 21 per hand keypoints respectively from the COCO-Wholebody dataset~\cite{jin2020whole}, and 9 spine annotations from SpineTrack~\cite{khan2025towards} are used. Across the five experiences, the number of keypoints changes from 17, 23, 91, and 133 to 142 in the last experience. Training images from COCO dataset, shared by all three datasets, rule out any domain shifts. This allows the benchmark to solely assess skeleton growth capabilities of different continual learning strategies. As in the domain-incremental track, an off-the-shelf body pose estimator is used for the first experience, and the other four experiences are fine-tuned with maximum 10 epochs per experience.

Unlike the domain-incremental scenario, the model architecture must adapt to accommodate increasing number of keypoints. In regularization-based CL methods, this adaptation is limited to the head layers only, which are expanded to output the required number of channels while preserving learned weights of the existing channels. This setting tests the model's ability to extend its output space over time without losing earlier keypoint accuracy—a crucial requirement in custom-skeleton scenarios. Evaluation of this proposed class-incremental benchmark is left as a future work to maintain focus on domain shifts in this work.

\subsection{Discussion}
Across all domain-incremental tracks, the constrained 1k/10-epoch budget makes naïve FT highly unstable: it adapts strongly to the current domain but rapidly erodes earlier ones, often falling below the frozen reference model. The experience–experience matrices for FT in \cref{fig:heatmap-density,fig:heatmap-lighting,fig:heatmap-mode} make this clear—off-diagonal entries fade fast, especially under Lighting and Modality, where forgetting compounds after each shift.

Among regularizers, \textbf{LFL} provides the most reliable stability under photometric changes. In both the tables and the heatmaps, its diagonals remain comparatively high across WL$\rightarrow$LL$\rightarrow$VLL$\rightarrow$ELL, and its off-diagonals decay slowest. \textbf{LwF} handles mild shifts well and achieves the best single-step Depth AP, but its sequential matrices reveal larger cumulative drift. \textbf{EWC} retains earlier domains better in the Modality sequence, yet its diagonals under strong shifts are consistently lower, indicating limited plasticity.

The relative difficulty of the three benchmarks is also visible in the heatmaps: Density shows the mildest degradation, Lighting induces intermediate drift, and Modality exhibits the steepest collapse—particularly once Depth is introduced. The persistent RGB$\rightarrow$Depth gap across all methods underscores that regularization alone is insufficient for cross-sensor adaptation. Overall, PoseAdapt exposes clear stability–plasticity trade-offs that become increasingly pronounced under severe distribution shifts.

\section{Conclusion}
\label{sec:conclusion}

We presented \textbf{PoseAdapt}, a standardized benchmark and toolkit for continual human pose estimation under controlled conditions: fixed lightweight backbones, no access to past data, and tightly bounded adaptation budgets. Evaluating FT, EWC, LFL, and LwF across density, lighting, and modality shifts, reveals consistent patterns:  
(1) FT is brittle and frequently underperforms the frozen reference;
(2) regularization improves retention but is sensitive to shift severity;
(3) LFL is the most stable across photometric domains, whereas LwF offers superior target-domain plasticity; and
(4) none of the methods handle RGB$\rightarrow$Depth robustly.

By unifying protocols, metrics (RA, AF), and shift generation pipelines, PoseAdapt establishes a reproducible foundation for studying sustainable adaptation in pose estimation.

\noindent \textbf{Impact.}
PoseAdapt provides a modular and reproducible framework that highlights concrete design targets for continual pose estimation: stronger feature alignment for modality changes, more stable regularizers for photometric shifts, and principled head-expansion strategies for skeleton growth. We hope the benchmark accelerates progress toward continual models suitable for long-term, real-world deployment.

\noindent \textbf{Limitations.}
Most shifts are synthetic, and while they isolate controllable factors, they do not fully capture real-world sensor characteristics or motion-induced artefacts. The fixed-backbone assumption focuses the evaluation on adaptation strategies but excludes architectural innovations that may improve robustness to severe shifts. Finally, the benchmark considers only 2D single-frame keypoints and does not address temporal consistency or 3D pose estimation. Moreover, while skeleton growth scenarios are supported, they are not explicitly benchmarked here. Instead, this work focused on presenting the PoseAdapt framework and a systematic benchmark under high-stress conditions. Performance of continual pose estimation on a real-world application where domain and keypoint shifts both appear together is also not investigated.

\noindent \textbf{Future Work.}
Future extensions should incorporate adapter-based or replay-informed continual learning, explore architectures with explicit cross-modal priors, and expand to transformers or bottom-up approaches. Extending PoseAdapt to video-based and 3D settings would allow systematic evaluation under temporal and geometric shifts. Broader domain coverage beyond synthetic shifts, such as clinical images or sports datasets, would further improve ecological validity.

\paragraph*{Acknowledgement.}
The work leading to this publication was co-funded by the European Union’s Horizon Europe research and innovation programme under Grant Agreement No 101135724 (project LUMINOUS) and Grant Agreement No 101092889 (project SHARESPACE).

{
    \small
    \bibliographystyle{ieeenat_fullname}
    \bibliography{main}
}

\end{document}